\documentclass[10pt,letterpaper]{article}

\usepackage{lineno}
\usepackage[top=0.85in,left=2.75in,footskip=0.75in,marginparwidth=2in]{geometry}
\usepackage[export]{adjustbox}
\usepackage[utf8]{inputenc}
\usepackage{pdfpages}

\usepackage{cite}
\usepackage{lmodern}
\usepackage{newtxtext,newtxmath}
\usepackage{nameref,hyperref}

\usepackage{microtype}
\DisableLigatures[f]{encoding = *, family = * }

\usepackage{indentfirst,csquotes}

\usepackage{changepage}

\usepackage[aboveskip=1pt,labelfont=bf,labelsep=period,singlelinecheck=off]{caption}

\makeatletter
\renewcommand{\@biblabel}[1]{\quad#1.}
\makeatother

\usepackage{lastpage,fancyhdr,graphicx}
\usepackage{epstopdf}
\usepackage{amsmath}
\fancyheadoffset[L]{2.25in}
\fancyfootoffset[L]{2.25in}

\usepackage{color}

\definecolor{Gray}{gray}{.25}

\usepackage{graphicx}

\usepackage{sidecap}

\usepackage{wrapfig}
\usepackage[fulladjust]{marginnote}
\reversemarginpar

\begin{document}
\vspace*{0.35in}

\begin{flushleft}
{\Large
\textbf{Generative QoE Modeling: A Lightweight Approach for Telecom Networks}
}


Vinti Nayar\textsuperscript{1},
Kanica Sachdev\textsuperscript{1},
Brejesh Lall\textsuperscript{1},
\\
\bigskip
\bf{1} IIT Delhi
\\

\bigskip
* Brejesh.Lall@ee.iitd.ac.in

\end{flushleft}

\section*{Abstract}
\noindent Quality of Experience (QoE) prediction plays a crucial role in optimizing resource management and enhancing user satisfaction across both telecommunication and over-the-top (OTT) services. While recent advances predominantly rely on deep learning models, this study introduces a lightweight generative modeling framework that balances computational efficiency, interpretability, and predictive accuracy. By validating the use of Vector Quantization (VQ) as a preprocessing technique, continuous network features are effectively transformed into discrete categorical symbols, enabling integration with a Hidden Markov Model (HMM) for temporal sequence modeling. This VQ-HMM pipeline enhances the model’s capacity to capture dynamic QoE patterns while supporting probabilistic inference on new and unseen data. Experimental results on publicly available time-series datasets—incorporating both objective indicators and subjective QoE scores—demonstrate the viability of this approach in real-time and resource-constrained environments, where inference latency is also critical. The framework offers a scalable alternative to complex deep learning methods, particularly in scenarios with limited computational resources or where latency constraints are critical.


\section{Introduction}
\begin{figure*}[t]
    \centering
    \includegraphics[width=\textwidth]{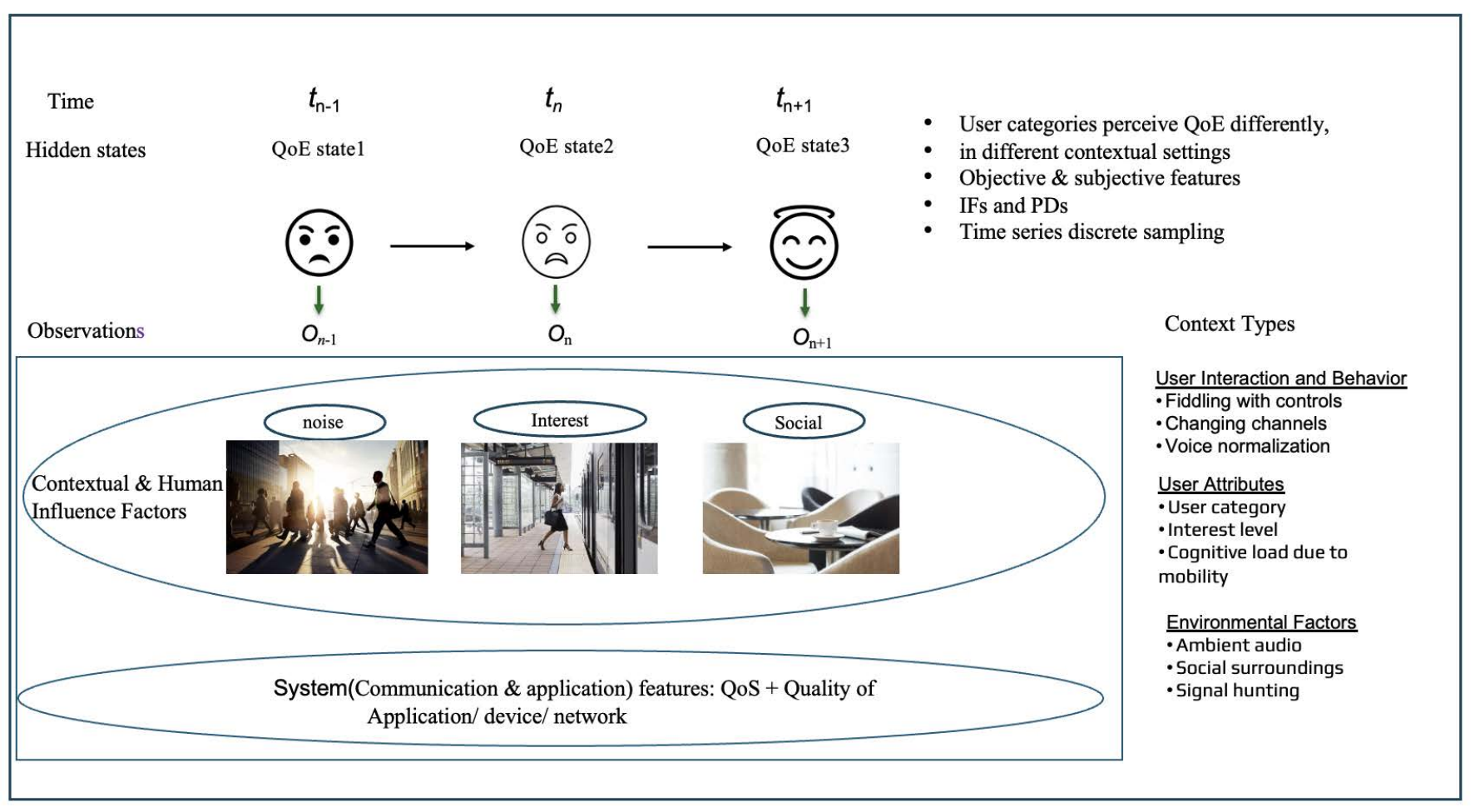} 
    \caption{QoE stack}
    \label{fig:QoE}
     
\end{figure*}

\noindent User expectations for service quality have grown steadily with the advent of new applications and services. As future telecom networks incorporate heterogeneous technologies across diverse deployment environments, ensuring Quality of Service (QoS) remains a central concern for network providers. QoE, a more evolved and user-centric concept, expands upon QoS by capturing a customer's perceived satisfaction or dissatisfaction with the service. It reflects an end-to-end evaluation of the user’s experience, involving many subjective elements that are still being uncovered, as noted in \cite{qualinet2013}.  
\newline \noindent Crucially, QoE unfolds over time—an aspect of paramount importance in its assessment, as emphasized in \cite{amour2015building}, \cite{amour2015hierarchical}, \cite{laghari2012toward}, \cite{husic2022multidimensional}.
The study in \cite{panahi2024machinelearningdrivenopensourceframework} argues against universal modeling strategies and advocates for user-centric, context-aware machine learning approaches to predict QoE more effectively. With the anticipated rise of immersive services, the ability to assess QoE in real time and close proximity to the user will become increasingly important \cite{schatz2013packets}. Various considerations for QoE assessment, including the use of complex time-series models with high predictive accuracy  \cite{10.1145/3673422.3674892}, \cite{panahi2024machinelearningdrivenopensourceframework}, \cite{eswara2019streaming} continue to be explored in a range of application scenarios. 
\newline \noindent QoE is shaped by a complex interaction between objective metrics and subjective dimensions. Figure~\ref{fig:QoE} illustrates the QoE stack. This figure illustrates the factors influencing QoE, including system-level, contextual, and human factors. Although this study recognizes the role of human, network, application, and contextual variables, it focuses primarily on system level modeling due to the limited availability of datasets that include Perceptual Dimensions (PDs).

\noindent While deep learning models have shown promise, their computational complexity and data requirements often limit their applicability in real-time or resource-constrained environments.  The experiments evaluate different modeling techniques on publicly available time-series datasets that incorporate both objective and subjective metrics. Through comparative analysis with other models, we highlight the trade-offs between prediction accuracy and computational latency, underscoring the practical advantages in telecom environments. 
\newline \noindent A key contribution of this work is the validation of Vector Quantization (VQ) as an effective preprocessing technique for transforming continuous valued network features into discrete categorical symbols, enabling their integration into sequence modeling frameworks for QoE estimation. This VQ-HMM pipeline enhances the model's ability to capture temporal QoE dynamics, resulting in significantly improved prediction accuracy. Furthermore, the approach has potential for applicability in No Reference (NR) and Near Reference (NRR) scenarios — conditions where direct ground truth for user QoE is unavailable. The generative structure of HMMs enables probabilistic inference on unseen user data, semi-supervised learning, and synthetic QoE sequence generation, making this approach well suited for deployment on lightweight edge platforms and wearable devices. To the best of our knowledge, this is among the first works to apply a VQ-driven HMM framework that demonstrates low latency QoE estimation on publicly available datasets, which can also support data augmentation in resource constrained environments.

\section{Related Works}
\noindent QoE estimation relies on a combination of objective features—which serve as quantifiable, reproducible indicators of QoS and subjective PDs, which are crucial for capturing user-centric experiential nuances \cite{amour2015building}, \cite{amour2015hierarchical}, \cite{bampis2017study}. Many existing studies utilize datasets that conform to ITU-T P.10/G.100 recommendations \cite{rec2017p}, with the Absolute Category Rating (ACR) scale commonly employed for annotating QoE. This standardized annotation facilitates consistent benchmarking across different models and studies. While recent QoE modeling efforts have been dominated by deep learning frameworks—both discriminative and generative—foundational time-series generative models like HMMs remain significantly underutilized. This is especially true in contexts where subjectively captured, real world datasets are available. 
One of the early attempts at applying HMMs to QoE was presented in  \cite{mitra2012qoe}, where the model was used to predict QoE during simulated handoffs in Heterogeneous Access Networks (HAN), relying on a small set of network level features guided by the ITU-T estimation framework. Similarly \cite{gandotra2017sector} proposed an adaptive resource allocation strategy leveraging HMMs, albeit evaluated on synthetic datasets. While these studies affirm the theoretical potential of HMM based modeling, they do not validate the approach on real world, user perceived QoE data, nor do they provide publicly accessible datasets, to the best of our knowledge, for further benchmarking. Thus limiting their practical applicability and reproducibility.

\medskip \noindent A comprehensive survey by \cite{kougioumtzidis2022survey} reviews the state-of-the-art in QoE modeling and underscores the importance of real time PDs in estimating user experience. While the dataset in \cite{Porcu2020EstimationOT} provides a rich set of subjective dimensions, it is limited to retrospective QoE annotations, collected at the end of user sessions. Likewise, \cite{amour2015building}, \cite{amour2015hierarchical} investigate different datasets with significant PD coverage, but again, only for post session assessments. The broader challenge of integrating human-centric features into QoE frameworks is addressed in \cite{barakovic2020quality}, which advocates for a 360 degree unified view encompassing both subjective quantification and integrated modeling. Furthermore, the limitations of subjective testing—including cost, scalability, and predictive generalizability—are highlighted in \cite{chen2014qos}. 
\newline \noindent In contrast to the simulated data approaches, datasets such as those in \cite{DBLP:journals/corr/BampisB17}, \cite{eswara2018streamingvideoqoemodeling} and \cite{bampis2017study} offer real time QoE annotations across diverse viewing scenarios. These datasets have been extensively modeled  but the application of HMMs to these data remains limited. Despite this gap in QoE specific literature, HMMs, have seen widespread application in other domains—most notably in Natural Language Processing (NLP) and genomics, where their ability to model sequential and temporal dependencies has been well validated in number of works \cite{hsu2010effects} and alike. \newline \noindent Table.~\ref{table1} provides a brief overview of the key methodologies and strategies used in the selected studies mentioned in the earlier paragraphs. Table.~\ref{table2} summarizes the comparison of related work.

\vspace{-.5cm}
\begin{table*}[p]
\caption{Related Works}
\label{table1}
\scriptsize
\begin{tabular}
{|p{2cm}|p{1cm}|p{1.5cm}|p{1.5cm}|p{1.5cm}|p{4cm}|}
\hline
\hline
\textbf{Paper Title/ Author} & \textbf{Dataset available} & \textbf{Methodology Used}  &  \textbf{Significant Findings} & \textbf{Practical Applications} & \textbf{Limitations}\\ 
\hline
 QoE Estimation and Prediction using Hidden Markov Models in Heterogeneous Access Networks / K Mitra & No & HMM  & HMM can accurately estimate QoE in handoffs in HANs (using the ITU-T E-model) & Application in Handoff protocols in HAN & 1. Limited in capturing nuances of user perception, as it assumes fixed mappings between network metrics with Objectively estimated QoE.  2. Simulates two  observation variables (One Way Delay, Round Trip Time) with Gaussian distribution \\
\hline
Context aware QoE modelling, measurement, and prediction in mobile computing systems / K. Mitra  & No & Naïve Bayes, Gaussian Naïve Bayes, Hybrid Bayesian Networks & CaQoEM efficiently learns and predicts QoE using BNs & Context aware approach called CaQoEM for QoE modelling. Novel introduction to QoE understanding & Non temporal \\
\hline
Study of temporal effects on subjective video quality of experience / Bampis & Yes & Correlation & Significant dataset for accuracy studies on temporal subjective QoE & Streaming  & Incorporating additional perceptual dimensions into observations can significantly enhance accuracy  \\
\hline
A hierarchical classification model of QoE Influence Factors/ Amour & Yes & Deterministic & Subjective perceptual dimensions are of significant importance & Importance of relevant IFs & Subjective QoE but only in retrospect \\
\hline
Streaming video qoe modeling and prediction: A long short term memory approach/ Eswara & Yes & LSTM & Enhanced Temporal accuracy.  Significant dataset for accuracy studies on temporal subjective QoE & Streaming video in telecom networks & 1. Challenge in resource limited deployments. 2. Requires incorporating additional perceptual dimension observations to increase accuracy in QoE.  \\
\hline
This proposed work & Yes & HMM & Transformation of multivariate continuous data enhancing temporal accuracy & Effective data transformation and validation of light weight generative model under NR and NRR conditions for subjective QoE & QoE is the sole subjective feature (PD). Requires incorporating additional perceptual dimension observations can significantly enhance accuracy  \\
\hline
\end{tabular}


\end{table*}

\begin{table*}[t]
\caption{Summarized comparison of Related studies}
\label{table2}
\scriptsize
\begin{tabular}
{|p{3cm}|p{3cm}|p{1.5cm}|p{1.5cm}|p{1.5cm}|}
\hline
\hline
\textbf{Study} & \textbf{Dataset Type} & \textbf{HMM usage}  &  \textbf{Realworld validation} & \textbf{Public Access}\\ 
\hline
Mira et al {\cite{mitra2012qoe}} & Simulated & QoE prediction  & \textbf{X} & \textbf{X} \\
\hline
Gandotra et al \cite{gandotra2017sector} & Synthetic & QoE prediction & \textbf{X} & \textbf{X} \\
\hline
Kougioumtzidis et al \cite{kougioumtzidis2022survey} & Survey & \textbf{X} & \textbf{-} & \textbf{-} \\
\hline
Porcu et al \cite{Porcu2020EstimationOT} & Retrospective Subjective & \textbf{X} & \checkmark & \checkmark \\
\hline
Amour et al \cite{amour2015building} \cite{amour2015hierarchical} & Retrospective PD & \textbf{X} & \checkmark & \checkmark\\
\hline
Bampis et al \cite{bampis2017study}, Eswara et al \cite{eswara2018streamingvideoqoemodeling} \cite{eswara2019streaming} & Real Time subjective & \textbf{X} & \checkmark & \checkmark  \\
\hline
\end{tabular}


\end{table*}
\section{Methodology}\label{sec3}
\noindent This section outlines the modeling framework used for QoE prediction, including its theoretical underpinnings, design rationale, architectural components, and evaluation setup. The proposed approach integrates a VQ step with a HMM to capture temporal QoE dynamics in a lightweight and interpretable manner. The following subsections detail the generative modeling foundation, the justification for using HMMs, the system architecture, and the datasets and evaluation strategy employed.
 \subsection{Theoretical Foundations}

\noindent Deterministic modeling and generative modeling represent two different approaches in the field of statistical modeling and machine learning. Deterministic models, such as linear regression and support vector machines, aim to learn the conditional probability  \( P(Y \mid X) \), , where X denotes observed features and Y the target variables. These models are designed to map inputs to outputs based on fixed rules or discriminative learning.

\noindent In contrast, generative models aim to learn the joint probability distribution. 

\[
P(X, Y) = P(Y) \cdot P(X \mid Y)
\]
\newline \noindent This allows for probabilistic inference, the generation of synthetic samples, and support for semi-supervised learning, as discussed in literature like \cite{bishop2006pattern}. Unlike deterministic models, generative approaches can capture hidden structure in the data and model relationships even in the presence of missing information.
\newline \noindent Here, \( P(Y) \) models the prior distribution over output classes or states, while \( P(X \mid Y) \) models the likelihood of observing the input given a specific output. Once the joint distribution is learned, the model can generate new, plausible examples \( (X', Y') \) by sampling from \( P(X, Y) \).

\noindent This generative capability is especially powerful in sequence modeling tasks, such as \textit{QoE estimation over time}, where temporal dependencies and latent state transitions can be effectively captured using models.
Within the context of  QoE estimation, which evolves over time and is often influenced by latent factors, generative modeling provides a compelling framework. In particular, generative models are well suited to tasks that require temporal reasoning, structured sequence prediction, and inference on incomplete or partially observed data. Their capacity to generate synthetic samples and support probabilistic interpretation aligns well with the challenges of modeling user experience in dynamic and context dependent environments.

\subsection{Model justification: Why HMMs?}
\noindent HMMs offer a structured and efficient way to model sequential data in the presence of hidden states. They are particularly effective when the observed data is generated by underlying latent processes that evolve over time, making them highly relevant for QoE estimation.
QoE is not only time dependent but also influenced by latent perceptual states that are not directly observable. HMMs can capture this by modeling the probabilistic transitions between hidden states and the emission of observable outputs from each state. This structure allows HMMs to infer the most likely underlying sequence of user experience states based on noisy or incomplete observations—a valuable capability in realistic deployment scenarios.
\newline \noindent Compared to deep learning models, which often require large volumes of training data and computational resources, HMMs are lightweight, interpretable, and easier to deploy in real time or edge environments. Their transparency enables a better understanding of state transitions and the influence of input features on perceived QoE. Additionally, the use of HMMs facilitates probabilistic reasoning, temporal alignment, and generation of synthetic data, all of which are essential for QoE modeling under NR and NRR conditions.

\subsection{System Architecture and workflow}

\noindent The proposed QoE prediction pipeline begins with a preprocessing stage in which multivariate network observations are mapped to latent user experience states. This is achieved through VQ, which transforms continuous valued input features into discrete categorical tokens adopting K means clustering algorithms. Each token represents a quantized version of the original observation, enabling efficient modeling with a discrete state sequence framework.
These quantized vectors are then passed into a first order HMM, which is trained to infer the underlying hidden states based on observed token sequences. The training process involves estimating the state transition matrix, the observation emission matrix, and the initial state probabilities from labeled sequences. Inference on new, unlabeled sequences is performed using the Viterbi algorithm, which computes the most probable sequence of hidden states corresponding to the observed input.
\newline \noindent This architecture supports both supervised training and unsupervised inference, making it adaptable to various deployment conditions. It is particularly effective in scenarios with limited or noisy ground truth, as the probabilistic structure of the model enables robust learning and interpretation. The integration of VQ and HMM results in a compact and interpretable pipeline that is well suited for real time and resource constrained environments.

\noindent 
Summarizing the two phases of the deployment into Training and Inference phase:

\textit{Training Phase}
\begin{itemize}
    \item Input: Labeled sequences (observations + corresponding states)
    \item Estimate transition probabilities $A$
    \item Estimate emission probabilities $B$ 
    \item Estimate initial distribution $\pi$
\end{itemize}

\textit{Inference Phase}
\begin{itemize}
    \item Input: Unseen sequence of multivariate observations
    \item Apply Viterbi algorithm to find the most probable state sequence
\end{itemize}

\subsection{Dataset and Experimental setup}

\noindent Experiments were carried out on a variety of publicly available datasets \cite{{Porcu2020EstimationOT}, {bampis2017study}, {eswara2019streaming}, {amour2015building}}. 
\noindent The datasets referred in the proposed work are selected for:
\begin{enumerate}
        \item Time sequence series
        \item Multivariate  observations
        \item Subjective QoE evaluated by the user
        
\end{enumerate}
\noindent The sample results presented in this section are derived from datasets generated through controlled experiments in \cite{eswara2018streamingvideoqoemodeling}, incorporating six multivariate network based predictors. The input features consist of multivariate sequences derived from raw telemetry data, including parameters such as network throughput, latency, and packet loss. These continuous valued features are discretized through VQ, which maps each data point to one of a predefined set of discrete tokens. The quantization step not only reduces computational complexity but also facilitates the application of discrete state sequence modeling via HMMs.

\noindent These predictors are grounded in established Video Quality Assessment (VQA) metrics, like Multi Scale Structural Similarity Index (MSSSIM) and Peak Signal-to-Noise Ratio (PSNR). The perceived QoE was captured using a continuous user reported rating scale ranging from 1 to 100. Among the predictors, MSSSIM and PSNR exhibited the highest contribution to prediction accuracy. While the other system parameters had comparatively lower individual impact, however their contributions were significant in enhancing overall QoE estimation.

\begin{figure*}[t]
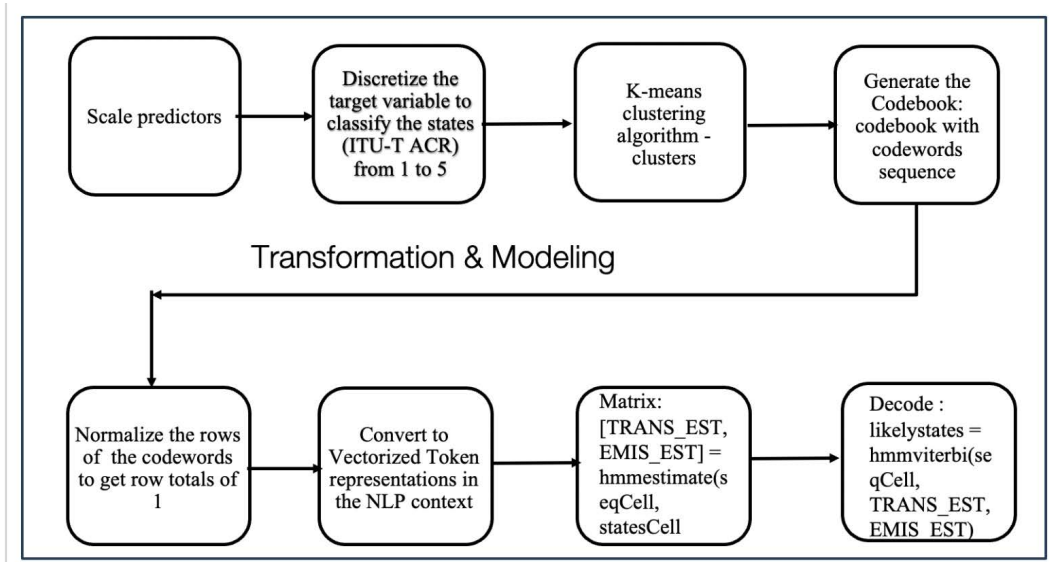

    \centering
     \resizebox{140mm}{!} {\includegraphics *{BoW5.pdf}}
    \caption{Transformation and Modeling pipeline}
    \label{fig:transform}
     
\end{figure*}

\noindent The  Transformation and modeling method employed are summarized in Figure~\ref{fig:transform}. This illustrates an overview of the workflow where multivariate network data is quantized using VQ and modeled using HMM for QoE prediction.

\subsection{Limitations}

\noindent A critical limitation in this domain is the scarcity of publicly available datasets that incorporate context aware or human centric features in real time settings. Consequently, this study relied on datasets emphasizing system level (network) metrics. Although system features are not the sole determinants of user QoE, they exert a substantial influence and enable reasonably accurate predictive modeling. Prior literature demonstrates that several advanced modeling techniques have successfully achieved high prediction accuracy based solely on such system parameters. However, the absence of comprehensive, multi dimensional datasets especially those integrating user context, content characteristics, and perceptual feedback continues to pose a challenge to the development of holistic, user centered QoE models.

\section{Results and Discussions}\label{sec}
\subsection{Accuracy and Latency Trade-offs}

\vspace{-10pt}
\begin{figure*}[htpb]
  \centering
  \resizebox{160mm}{!} {\includegraphics{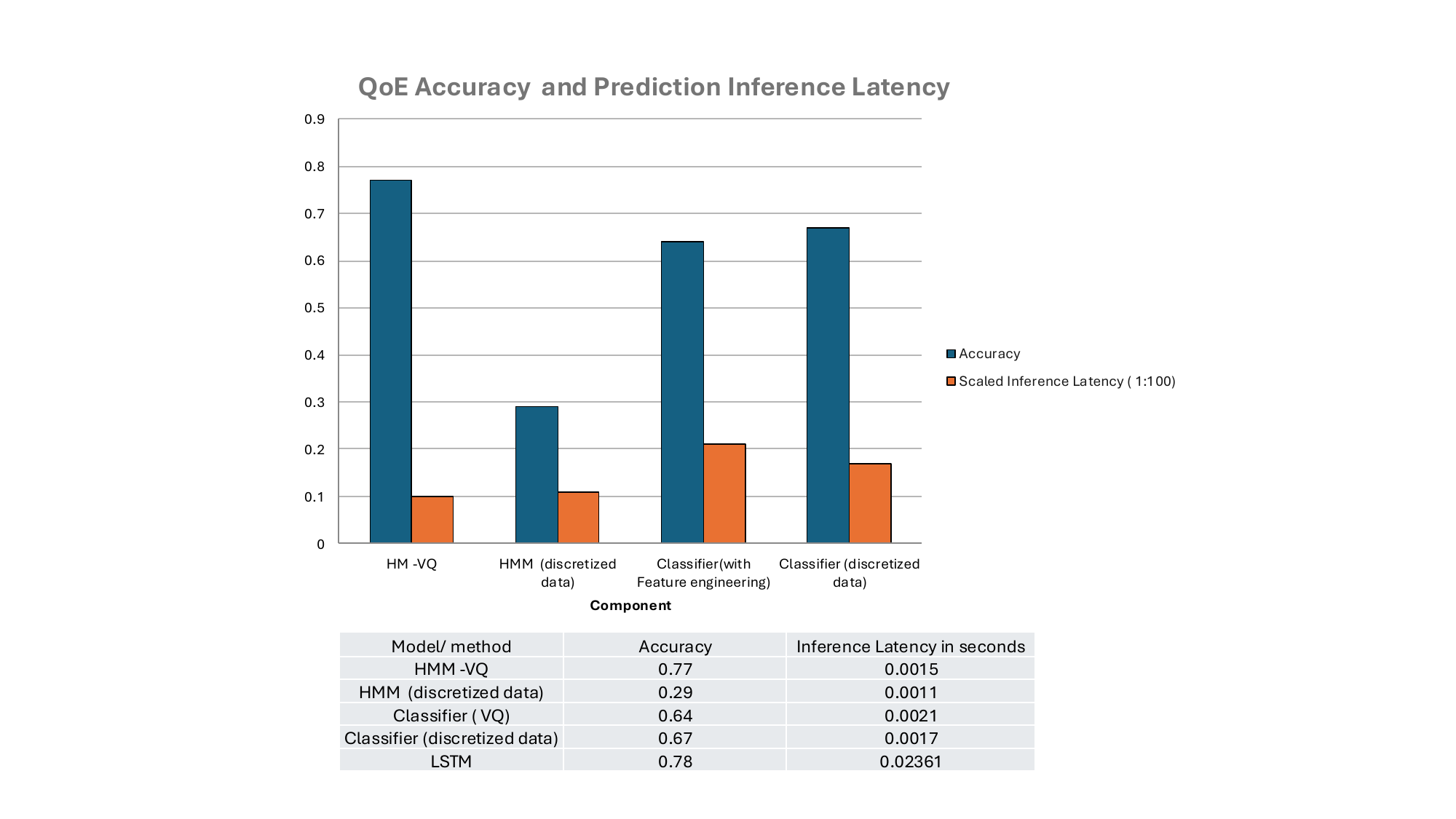}}
  \caption{Comparative analysis of models in terms of accuracy and latency}
  \label{fig:models}
\end{figure*}
\vspace{1cm}

 
\noindent Figure~\ref{fig:models} presents a comparative evaluation of multiple models for QoE prediction, focusing on both accuracy and prediction inference latency. The models assessed include HMMs, Long Short Term Memory networks (LSTMs), Generative Adversarial Networks (GANs), and Variational Autoencoders (VAEs). Notably, GANs and VAEs exhibited higher latency, which is not shown in the figure.
\newline\noindent HMM-VQ demonstrates the highest prediction accuracy of 0.77, paired with a minimal inference latency of 0.0015 seconds, making it highly suitable for real time QoE prediction. The integration of domain specific feature engineering significantly enhances the model’s ability to capture temporal QoE variations, providing an optimal balance between predictive accuracy and computational efficiency.
\noindent In contrast, the HMM with binned data yields suboptimal performance, with an accuracy of 0.29, despite its low inference latency of 0.0011 seconds. This substantial degradation in accuracy highlights the potential drawbacks of scalar discretization, which likely leads to the loss of critical temporal and contextual information, thereby negatively affecting the model's predictive power.
\newline \noindent Classifier models exhibit a range of performance outcomes. The classifier with feature engineering attains an accuracy of 0.64 and a latency of 0.0021 seconds, while the classifier employing discretized data achieves a marginally higher accuracy of 0.67, accompanied by a reduced latency of 0.0017 seconds. These results suggest that discretization, in this specific context, may simplify the feature space and enhance the classifier’s performance by reducing complexity.
\newline \noindent The LSTM model, although capable of achieving higher accuracy through fine tuning, incurs a substantial inference latency of approximately 0.023 seconds, which is roughly 10 times greater than the baseline models. Advanced models such as LSTMs, GANs, and VAEs, while demonstrating strong representational power, introduce significant inference latency. These models are thus more suitable for offline processing or cloud based batch inference, as their high latency makes them impractical for latency sensitive applications such as edge computing or wearable devices.
\newline \noindent In summary, the HMM-VQ model stands out as the most effective solution for real time QoE prediction, offering both high accuracy and minimal latency, thereby making it ideal for deployment in resource constrained, real time environments.

 \subsection{Precision Recall Analysis}
\begin{figure*}[htpb]
   \centering
   \resizebox{160mm}{!} {\includegraphics{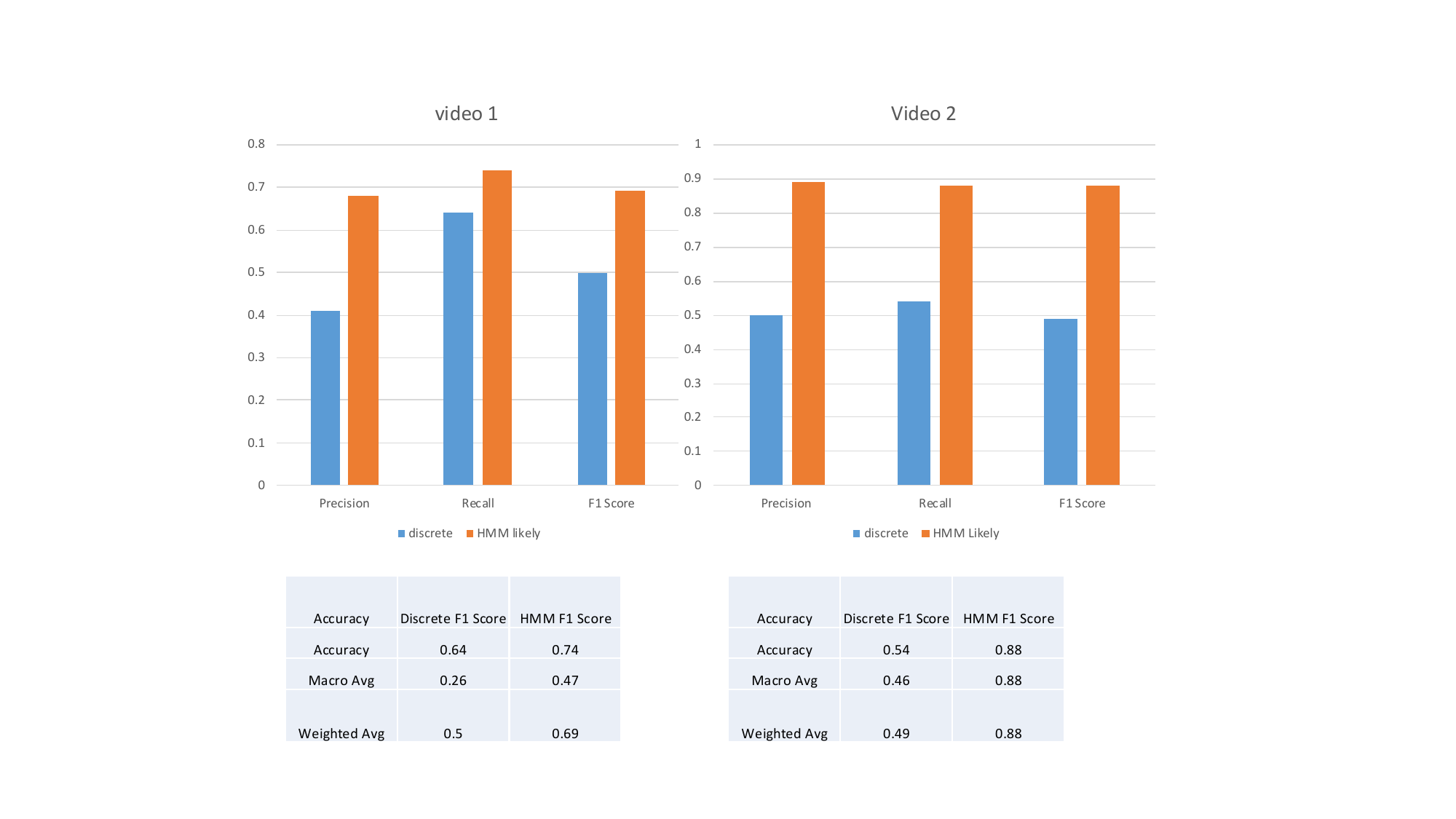}}
   \caption{ Precision-Recall Analysis}
   \label{fig:classifier}
 \end{figure*}

 \noindent  Fig.~\ref{fig:classifier} compares the performance of classifier based and HMM based models using the MATLAB toolkit across multiple video sequences. For Sequence 1, the classifier achieved a precision of 0.41, whereas the HMM derived "Likely" sequence improved this to 0.68, indicating a notable reduction in false positives. Recall also increased from 0.64 (classifier) to 0.74 (HMM), and the F1 score increased from 0.50 to 0.69, reflecting a more effective balance between precision and recall in the HMM based approach.

 \noindent 

\vspace{-10pt}
\begin{figure*}
 \begin{center}
    \resizebox{170mm}{!} {\includegraphics{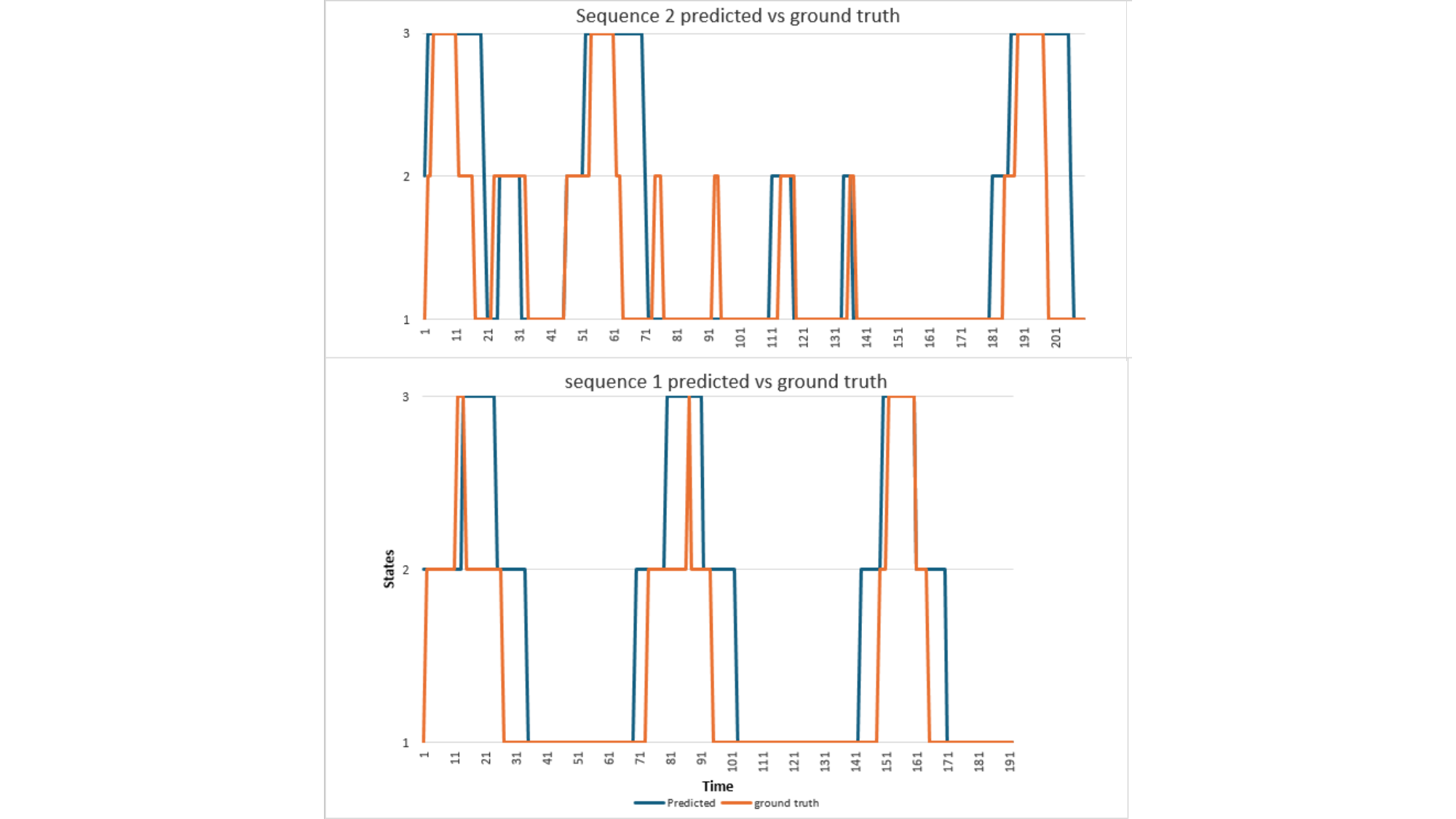}}
    \caption {Experimental Results in time domain }
  \label{fig:temporal graphs}
  \end{center}
\end{figure*}

\subsection{Capturing Temporal QoE Variations}
\noindent Figure~\ref{fig:temporal graphs} illustrates representative outcomes from the temporal variation analysis. Despite employing a first order HMM, the model effectively captures temporal dynamics in the QoE state sequence aligned with network-level observations. Except for transient deviations during state transitions (lasting only a few milliseconds), the inferred "Likely" state sequence closely matches the ground truth, as evidenced in the corresponding time series plots.

\noindent The generative structure of the HMM not only facilitates accurate hidden state inference but can also enable the generation of synthetic sequences for downstream training and evaluation. Leveraging VQ significantly reduces input dimensionality, ensuring computational efficiency suitable for latency sensitive edge deployments.

\noindent Crucially, the model’s interpretable architecture—through learned transition and emission probabilities—provides transparency into QoE dynamics, in contrast to black-box deep learning models. This combination of low latency inference, reduced complexity, and interpretability underscores the VQ-HMM framework's practicality for scalable QoE estimation in future telecom networks.

\section{Conclusions and Future Work}
\noindent This paper proposed a lightweight generative framework for QoE prediction, combining vector quantization and Hidden Markov Models (VQ-HMM) to enable efficient, interpretable sequence modeling. The framework effectively captures temporal QoE dynamics with high accuracy and low computational complexity, demonstrating suitability for real time, resource constrained applications.

\noindent Experimental results validated the model’s potential performance in both NRR and NR scenarios, particularly where subjective feedback is limited. In addition to predictive accuracy, the model offers transparency through interpretable state transitions, addressing key limitations of deep learning based methods.

\noindent Future work will incorporate perceptual dimensions to better align predictions with user-centric QoE factors, explore hybrid generative-discriminative models to enhance representational capacity, and expand real time annotated datasets to improve generalization across diverse deployment environments.

\nolinenumbers

\bibliography{library}

\bibliographystyle{abbrv}

\end{document}